\documentclass[letterpaper]{article} 
\usepackage{aaai2026}  
\usepackage{times}  
\usepackage{helvet}  
\usepackage{courier}  
\usepackage[hyphens]{url}  
\usepackage{graphicx} 
\usepackage{natbib}  
\usepackage{caption} 
\usepackage{algorithm}
\usepackage{algorithmic}
\usepackage{pifont} 
\usepackage{placeins}
\usepackage{booktabs}
\usepackage{graphicx}
\usepackage{multirow}
\usepackage{tabularx}
\usepackage{graphicx}
\usepackage{multirow}
\usepackage[utf8]{inputenc} 
\usepackage[T1]{fontenc}    
\usepackage{url}            
\usepackage{booktabs}       
\usepackage{amsfonts}       
\usepackage{nicefrac}       
\usepackage{microtype}      
\usepackage{xcolor}  
\usepackage{amsmath}
\usepackage{diagbox} 
  \usepackage{xcolor}
\usepackage{newfloat}
\usepackage{listings}
\DeclareCaptionStyle{ruled}{labelfont=normalfont,labelsep=colon,strut=off} 
\floatstyle{ruled}
\newfloat{listing}{tb}{lst}{}
\floatname{listing}{Listing}
\title{SynWeather: Weather Observation Data Synthesis across Multiple Regions and Variables via a General Diffusion Transformer}
\author{
    Kaiyi Xu\textsuperscript{\rm 1,\rm 2}\equalcontrib
    Junchao Gong\textsuperscript{\rm 2,\rm 3}\equalcontrib
    Zhiwang Zhou \textsuperscript{\rm 2,\rm 4}
    Zhangrui Li \textsuperscript{\rm 2,\rm 5}
    Yuandong Pu \textsuperscript{\rm 2,\rm 3}
    Yihao Liu \textsuperscript{\rm 2} \\
    Ben Fei \textsuperscript{\rm 2,\rm 6}
    Fenghua Ling \textsuperscript{\rm 2}
    Wenlong Zhang \textsuperscript{\rm 2}\thanks{Corresponding author.}
    Lei Bai \textsuperscript{\rm 2\textdagger}
}
\affiliations{
    \textsuperscript{\rm 1}University of Science and Technology of China
    \textsuperscript{\rm 2}Shanghai Artificial Intelligence Laboratory
    \textsuperscript{\rm 3}Shanghai Jiao Tong University
    \textsuperscript{\rm 4}Tongji University
    \textsuperscript{\rm 5}Nanjing University
    \textsuperscript{\rm 6}The Chinese University of Hong Kong

xukaiyi@pjlab.org.cn, gjchimself@sjtu.edu.cn,
zhouzhiwang@pjlab.org.cn,
652022280007@smail.nju.edu.cn,
puyuandong01061313@gmail.com,
liuyihao@pjlab.org.cn,
benfei@cuhk.edu.hk,
lingfenghua@pjlab.org.cn,
zhangwenlong@pjlab.org.cn,
bailei@pjlab.org.cn
}

\usepackage{bibentry}

\begin{document}

\maketitle

\begin{abstract}
With the advancement of meteorological instruments, abundant data has become available. 
However, due to instruments’ intrinsic limitations such as environmental sensitivity and orbital constraints, raw data often suffer from temporal or spatial gaps, making it urgent to leverage data synthesis techniques to fill in missing information. 
Current approaches are typically focus on single-variable, single-region tasks and primarily rely on deterministic modeling. 
This limits unified synthesis across variables and regions, overlooks cross-variable complementarity and often leads to over-smoothed results. 
To address above challenges, we introduce SynWeather, the first dataset designed for \textbf{Unified Multi-region and Multi-variable Weather Observation Data Synthesis}. 
SynWeather covers four representative regions: the Continental United States, Europe, East Asia, and Tropical Cyclone regions, as well as provides high-resolution observations of key weather variables, including Composite Radar Reflectivity, Hourly Precipitation, Visible Light, and Microwave Brightness Temperature. 
In addition, we introduce SynWeatherDiff, a general and probabilistic weather synthesis model built upon the Diffusion Transformer framework to address the over-smoothed problem. 
Experiments on the SynWeather dataset demonstrate the effectiveness of our network compared with both task-specific and general models. 
Moreover, SynWeatherDiff is able to generate results that are both fine-grained and accurate in high-value regions.
Through the dataset and baseline model, we aim to advance meteorological downstream tasks and promote the development of general models for weather variable synthesis.
\end{abstract}

{\small
\begin{links}
    \link{Website}{https://dtdtxuky.github.io/SynWeather-Proj/}
    \link{Code}{https://github.com/Dtdtxuky/SynWeather}
\end{links}
}


\begin{figure}[t]
    \centering
    \includegraphics[width=1.05\linewidth]{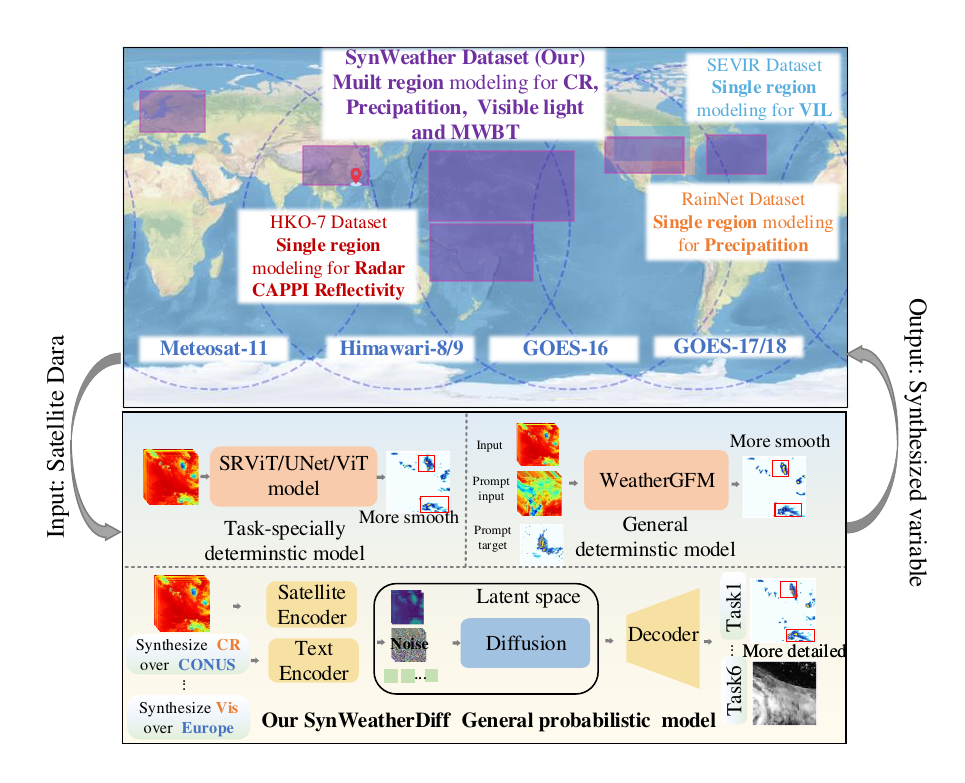}
    \caption{Overview of datasets and pipelines in weather variable synthesis. Compared to existing single-region, single-variable and deterministic modeling, SynWeather enables general multi-region, multi-variable probabilistic modeling.}
    \label{fig:5}
\end{figure}

\section{Introduction}
With the development of various meteorological observation instruments, including geostationary satellites, polar-orbiting satellites and radar systems, a wide range of weather data has become available~\cite{schmit2017closer, bessho2016introduction}. Such data have contributed to advancements in weather forecasting, disaster monitoring, and climate research~\cite{bauer2015quiet, joyce2009review, stephens2002cloudsat} and have been used to form various datasets~\cite{zhou2025scientists, zhao2025msearth, wang2025omniearth}. However, due to the intrinsic characteristics and deployment constraints of the aforementioned instruments, original weather data face limitations in both temporal and spatial coverage. For instance, radar observations often suffer from sparse coverage in regions with complex terrain or limited economic development~\cite{germann2022weather, ovchynnykova2025assessing}. Similarly, visible satellite images are also unavailable at night~\cite{harder2020nightvision, pasillas2024turning}. 

To bridge these gaps in spatial and temporal coverage, weather variable synthesis has emerged as a significant area of research~\cite{oliver1990kriging, liu2023pristi, hayawi2025climate}. 
Recent methods have used satellite infrared and microwave observations to reconstruct Composite Reflectivity~\cite{stock2024srvit, he2025diffsr}, estimate precipitation fields~\cite{gorooh2022deep, cannon2024deep}, and generate visible light on night~\cite{harder2020nightvision, chirokova2023proxyvis}. 
\citeauthor{li2025generative} first used diffusion models conditioned on high-resolution geostationary infrared data to extend the spatial coverage of microwave observations. Apart from these task-specific models, WeatherGFM~\cite{zhao2024weathergfm} as a general model has also been applied to a variety of weather understanding tasks, including radar and visible light synthesis. 

\begin{figure*}[t]
    \centering
    \includegraphics[width=0.80\linewidth]{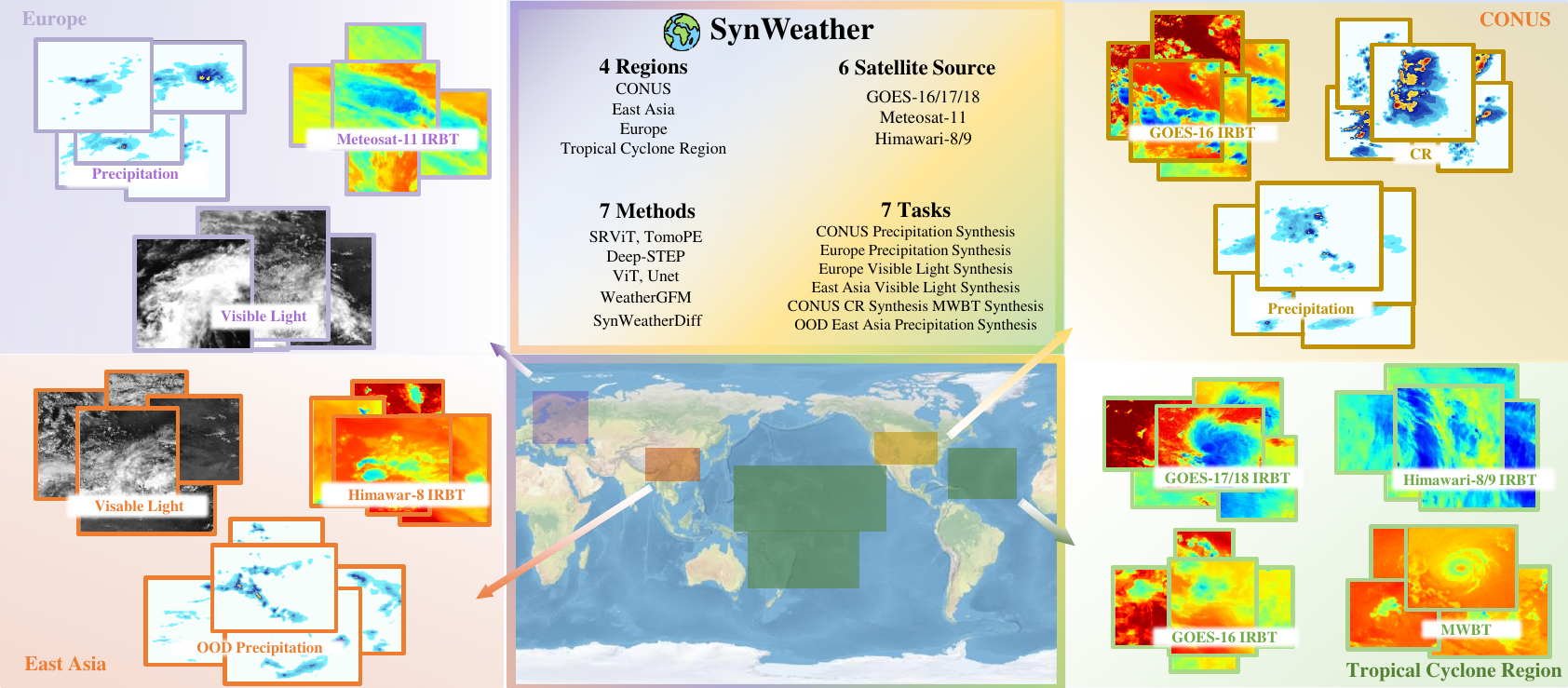}
    \caption{Overview of SynWeather. SynWeather is a comprehensive dataset that covers four distinct regions and four key weather observation variables, integrating data from six satellite sources as a condition to support seven synthesis tasks. Extensive evaluations are conducted on seven models, comprising both task-specific and general synthesis models.}
    \label{fig:1}
\end{figure*}

However, weather variable synthesis is still limited in the following aspects:
(1) \textit{Lack of Global-Scale and Multi-Variable Observations Dataset}: Most existing datasets for weather variable synthesis are limited to single-region or/and single-variable. For example, HKO-7~\cite{shi2017deep} and SEVIR~\cite{veillette2020sevir} focus solely on radar-related variables over Hong Kong and the Continental United States (CONUS), respectively. While some global datasets like RainBench~\cite{schroeder2020rainbench} include multiple weather variables, these are mostly derived from reanalysis rather than observation. Moreover, due to the limited geographical coverage, existing datasets rely on a single satellite source. For instance, SEVIR~\cite{veillette2020sevir} only uses three GOES satellites, and Digital Typhoon~\cite{kitamoto2023digital} only uses one Himawari channel. (2) \textit{Customized Deterministic Modeling}: Current weather variable synthesis methods use a special network for a particular weather variable synthesis task. Besides, these methods are often deterministic models trained with pixel-wise squared loss. As a result, the outputs are often overly smooth and unable to capture high-intensity areas during severe weather events. Moreover, different regions and observation variables can be viewed as multi-modalities reflecting the atmospheric state, but existing datasets and modeling overlook their cross-modal complementarities. In summary, weather variable synthesis faces an important challenge: \textbf{\textit{How to perform general-purpose generative modeling of multiple weather observation variables across multiple regions?}}

\begin{table*}[h]
\centering
{\scriptsize
\begin{tabular}{@{}ccccccccc@{}}
\toprule
\multirow{3}{*}{\diagbox{Region}{Data}} & \multicolumn{3}{c}{Input} & \multicolumn{3}{c}{Target} & \multirow{3}{*}{Sample numbers} & \multirow{3}{*}{Year} \\
\cmidrule(lr){2-4} \cmidrule(lr){5-7}
 & Satellite & Band & \begin{tabular}[c]{@{}c@{}}Spatial\\Res. (km)\end{tabular} & Variable & Source & \begin{tabular}[c]{@{}c@{}}Spatial\\Res. (km)\end{tabular} & \\
\midrule
\multirow{2}{*}{CONUS} & \multirow{2}{*}{GOES-16} & \multirow{2}{*}{C07-16} & \multirow{2}{*}{2} & CR & GREMLIN CONUS3 Dataset & 3 & 142k & \multirow{2}{*}{2020–2022} \\
 & & & & Precipitation & MRMS & 1 & 20k & \\
\midrule
\multirow{2}{*}{Europe} & \multirow{2}{*}{Meteosat-11} & \multirow{2}{*}{\begin{tabular}[c]{@{}l@{}}IR\_016-134\\ WV\_062-073\end{tabular}} & \multirow{2}{*}{3} & Visible light & Meteosat-11 & 3 & 372k & \multirow{2}{*}{2019–2021} \\
 & & & & Precipitation & EURADCLIM & 1 & 25k & \\
\midrule
\multirow{2}{*}{East Asia} & \multirow{2}{*}{Himawari-8} & \multirow{2}{*}{C07-16} & \multirow{2}{*}{2} & Visible light & Himawari-8 & 2 & 503k & 2019–2021 \\
 & & & & Precipitation & GPM & 10 & 15k & 2021.7 \\
\midrule
\multirow{2}{*}{\begin{tabular}[c]{@{}c@{}}TC\\Region\end{tabular}} & GOES-16/17/18 & C07-16 & \multirow{2}{*}{2} & \multirow{2}{*}{MWBT} & AMSR-2 & 7×12, 3×5 & \multirow{2}{*}{9k}& \multirow{2}{*}{2015–2023} \\
 & Himawari-8/9 & C07-16 & & & GMI & 8.6×14, 4.4×7.2 & \\
\bottomrule
\end{tabular}
}
\caption{Detailed information of SynWeather. Spatial resolution is denoted as``Spatial Res''}
\label{tab:dataset_description}
\end{table*}

To address the above challenge, we first introduce SynWeather, a dataset that supports unified multi-region and multi-variable weather variable synthesis. The dataset spans four key regions: the CONUS, Europe, East Asia, and Tropical Cyclone regions (TC regions), and integrates 10 channels infrared observations from corresponding global geostationary satellites (GOES-16/17/18, Meteosat-11, and Himawari-8/9) to synthesize a variety of weather variables including composite reflectivity (CR), precipitation, visible light, and microwave brightness temperature (MWBT). Building upon SynWeather, we design six standard tasks and one out-of-distribution (OOD) task to comprehensively evaluate weather variable synthesis models. We further propose SynWeatherDiff, a general-purpose weather synthesis model built on a diffusion transformer. It leverages text prompts to guide the generation of diverse weather variables across different regions and satellite sources. By adopting a probabilistic modeling framework, SynWeatherDiff is capable of generating fine-grained spatial structures and recovering high-intensity regions, which are often missed by deterministic models. Moreover, the synthesized results can serve as valuable inputs for both nowcasting~\cite{gong2024postcast, he2025radarqa, xu2025syncast} and medium-range forecasting systems~\cite{chen2025operational}, as well as data assimilation~\cite{sun2025align}.
Our main contributions can be summarized as:
\begin{itemize}
    \item We construct SynWeather, the first standardized dataset supporting unified multi-region and multi-variable weather observation data synthesis tasks.
    \item We propose SynWeatherDiff, the first generative and probabilistic framework that unifies many variables and regions under a single text-prompt-driven interface.
    \item Extensive experiments demonstrate the effectiveness of SynWeatherDiff over both task-specific and general models, particularly in generating fine-grained details. 
\end{itemize}

\section{SynWeather Dataset}

\subsection{Dataset Collection}
The input data are collected from all infrared channels of six geostationary satellites (i.e., GOES-16/17/18, Himawari-8/9, and Meteosat-11) and could seamlessly cover the four target regions as shown in Fig.~\ref{fig:1}. These satellites capture full-disk images every 10–15 minutes at spatial resolutions from 0.5 to 4 km. The target weather variables are sourced from authoritative datasets in each respective region: composite reflectivity (CR) from the GREMLIN CONUS3 dataset~\cite{hilburn2023gremlin} and 1-hour quantitative precipitation estimates (QPE) from Multi-Radar Multi-Sensor (MRMS) in the CONUS; radar gauge fused precipitation from EURADCLIM~\cite{overeem2023euradclim} in Europe; satellite-based precipitation from Global Precipitation Measurement (GPM) mission in East Asia; and microwave brightness temperatures (MWBT) from AMSR-2 and GMI provided by the Tropical Cyclone Precipitation, Infrared, Microwave, and Environmental Dataset (TC PRIMED)~\cite{razin2023tropical} for TC regions. Details of the data coverage and composition are summarized in Table~\ref{tab:dataset_description}.

\subsection{Dataset Processing}
Since the original inputs and targets vary temporally and spatially, all data are first standardized to a unified resolution of 1-hour and 4-kilometer and aligned by the closest timestamps. To ensure consistent input size for training, all regional data are cropped to $256 \times 256$ patches using a sliding window with a 128-pixel stride. However, not all patches contain meaningful signals; therefore, we set two thresholds $\gamma_1$, $\gamma_2$ and keep patches that contain connected components formed by pixels that exceed \(\gamma_1\) in value with sizes larger than \(\gamma_2\). Finally, to align variable ranges, we log-transform precipitation for its long-tailed distribution, followed by min-max normalization.
For other variables, we directly apply min-max normalization.

\subsection{Dataset Statistics}
Apart from the MWBT that is collected in an event-driven manner during tropical cyclones in TC PRIMED dataset~\cite{razin2023tropical}, all other weather variables in SynWeather are sampled continuously at fixed time intervals. As a result, the overall number of raw samples is relatively balanced across variables. However, because precipitation events are inherently sparse in both space and time, many samples contain little or no rainfall. After filtering out these samples, the number of valid precipitation patches becomes significantly smaller than that of other variables. Table~\ref{tab:dataset_description} includes the sample number of different weather variables in the SynWeather. It is evident that visible light constitutes the majority of samples, followed by CR, while precipitation and MWBT data are relatively less.

\begin{figure*}[h]
    \centering
    \includegraphics[width=0.85\linewidth]{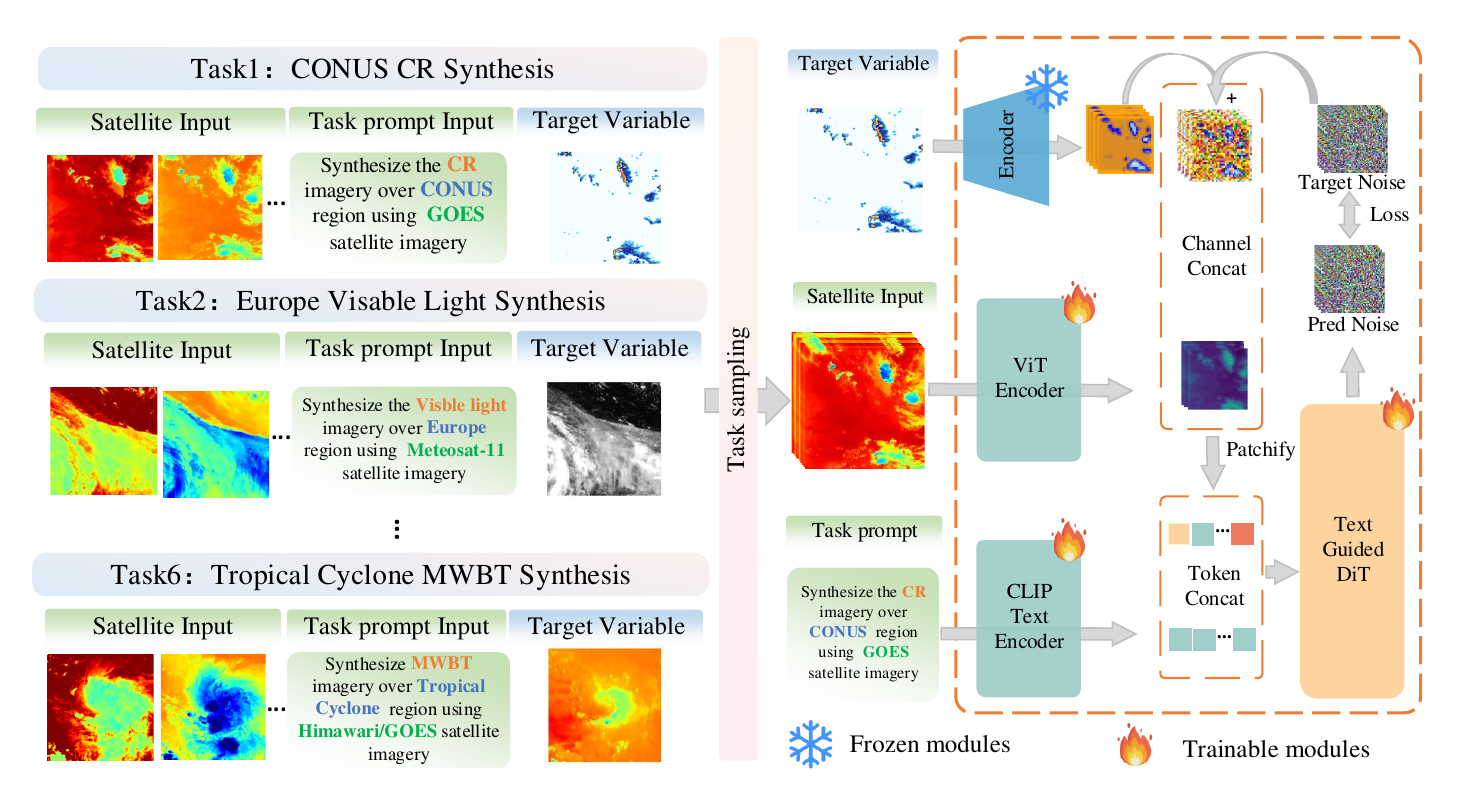}
    \caption{An overview of our SynWeatherDiff. The target variables are projected into a unified latent space using a general autoencoder. The satellite inputs are processed through a ViT-based encoder to extract features. A task-specific text prompt is encoded using a fine-tuned CLIP text encoder. The text tokens serve as conditional information to guide Text-Guided DiT for different weather synthesis tasks.}


    \label{fig:3}
\end{figure*}

\section{Proposed Baseline: SynWeatherDiff}
One major challenge in weather data synthesis is the diversity of satellite sources and observed variables across different regions and types. Under such conditions, a specialized modeling strategy requires training separate models for each region-variable pair, resulting in considerable computational and operational overhead. However, many weather variables are often physically correlated. For instance, precipitation can be inferred from CR using a Z-R relationship~\cite{wu2018dynamical, peng2022radar}, highlighting the potential for joint modeling across variables. Inspired by these observations, we propose SynWeatherDiff, a text-prompt-based general model that enables unified modeling and conditional generation across multiple regions and variables similar to unified models in natural images~\cite{chen2025exploring, pu2025lumina}.

\subsection{Problem Formulation}
Based on the SynWeather, we define seven weather observation variable synthesis tasks, including six standard tasks: (i) CR synthesis over CONUS, (ii, iii) precipitation synthesis over CONUS and Europe, (iv, v) visible light synthesis over East Asia and Europe, and (vi) MWBT synthesis over tropical cyclone regions, as well as a out-of-distribution task: precipitation synthesis over East Asia. 
For each task, we compare specialized models (trained on region- and variable-specific data) with a general model (trained on all regions and variables).
For single tasks, the specialized model \( f_{r,b} \) maps satellite observations \( X_r \in \mathbb{R}^{C \times H \times W} \) to weather observation variables \( Y_{r,b} \in \mathbb{R}^{1 \times H \times W} \):
\begin{equation}
Y_{r, b} = f_{r, b}(X_{r}),
\end{equation}
where $r$ denotes the target region, including CONUS, Europe (Eur), East Asia (EA), and TC regions. And $b$ represents the type of weather observation variable, containing CR, Precipitation (Prec), Visible Light (Vis) and MWBT; for OOD tasks, the specialized model $f_{r,b}$ generalizes as:
\begin{equation}
Y_{EA, Prec}^{OOD} = f_{r_1, Prec}(X_{EA}),
\end{equation}
where \(r_1 \in \{\text{CONUS}, \text{Eur}\}\).

In addition, we introduce a general model $g$ to perform all types of weather variable synthesis tasks, which takes both the satellite observation $X_r$ and text prompts $P_{r, b}$: ``\textit{Synthesize the $b$ variable over the $r$ region using corresponding satellite imagery. }'' as inputs to guide the generation of $Y_{r, b}$:
\begin{equation}
Y_{r, b} = g(X_{r}, P_{r, b}),
\end{equation}
Similarly, the OOD tasks are represented as:
\begin{equation} 
Y_{EA, Prec}^{OOD} = g(X_{EA}, P_{EA, Prec}).
\end{equation} 

\begin{table*}[h]
\centering
\small
\setlength{\tabcolsep}{4pt}
\renewcommand{\arraystretch}{1.2}
\begin{tabular*}{\textwidth}{@{\extracolsep{\fill}}cccccccccccc}
\toprule
\multirow{3}{*}{Task name} & \multicolumn{4}{c}{CR Synthesis} & \multicolumn{7}{c}{Precipitation Synthesis} \\
\cmidrule(lr{2pt}){2-5} \cmidrule(lr{2pt}){6-12}
& \multicolumn{4}{c}{CONUS} & \multicolumn{4}{c}{CONUS} & \multicolumn{3}{c}{Europe} \\
\cmidrule(lr){2-5} \cmidrule(lr){6-9} \cmidrule(l){10-12}
Metric & RMSE$\downarrow$ & CSI/25$\uparrow$ & CSI/35$\uparrow$ & CSI/40$\uparrow$ 
& RMSE$\downarrow$ & CSI/2$\uparrow$ & CSI/5$\uparrow$ & CSI/15$\uparrow$
& RMSE$\downarrow$ & CSI/2$\uparrow$ & CSI/5$\uparrow$ \\
\midrule
SRViT\textsuperscript{\#}               & 3.561 & 0.277 & 0.120 & 0.069 & -     & -     & -     & -     & -     & -     & -     \\
Deep-STEP\textsuperscript{\#}             & -     & -     & -     & -     & \textbf{0.916} & 0.262 & 0.111 & 0.007 & \underline{0.415} & \underline{0.083} & 0.016 \\
TomoPE\textsuperscript{\#}                & -     & -     & -     & -     & 0.986 & 0.247 & 0.149 & 0.036 & \textbf{0.413} & 0.060 & 0.009 \\
UNet\textsuperscript{\#}                  & 3.395 & 0.299 & 0.069 & 0.023 & 0.976 & 0.231 & 0.166 & 0.059 & 0.641 & 0.035 & 0.016 \\
ViT\textsuperscript{\#}                   & 3.487 & 0.309 & 0.141 & \underline{0.089} & 0.981 & 0.250 &0.157 & 0.038 & 0.497 & \underline{0.083} & \underline{0.044} \\
WeatherGFM\textsuperscript{\textdagger}   & \underline{3.124} & \underline{0.366} & \textbf{0.166} & 0.086 & 1.049  &  \underline{0.288} & \underline{0.198} & \underline{0.090} & 0.714     & 0.018 & 0.013 \\
SynWeatherDiff\textsuperscript{\textdagger}     & \textbf{2.820} & \textbf{0.382} & \underline{0.158} & \textbf{0.101} & \underline{0.976} & \textbf{0.312} & \textbf{0.223} & 
\textbf{0.113} & 0.569 & \textbf{0.084} & \textbf{0.079} \\
\toprule
\multirow{3}{*}{Task name} & \multicolumn{6}{c}{Visible Light Synthesis} & \multicolumn{5}{c}{MWBT Synthesis} \\
\cmidrule(lr){2-7}  \cmidrule(lr){8-12}
& \multicolumn{3}{c}{East Asia} & \multicolumn{3}{c}{Europe} & \multicolumn{5}{c}{Tropical Cyclone Region} \\
\cmidrule(lr){2-4} \cmidrule(lr){5-7} \cmidrule(l){8-12}
Metric & SSIM$\uparrow$   & PSNR$\uparrow$ & CSI/50$\uparrow$   & SSIM$\uparrow$ & PSNR$\uparrow$ & CSI/50$\uparrow$  & RMSE$\downarrow$ & SSIM$\uparrow$ & PSNR$\uparrow$  & LPIPS$\downarrow$ & CSI/300$\uparrow$\\
\midrule
ViT\textsuperscript{\#}    & \underline{0.870}     & \underline{20.87} & 0.672                 & 0.860    & \underline{24.03}       & 0.496            & \underline{4.768}          & 0.783          & 21.56          & 0.324          & \underline{0.792}          \\
UNet\textsuperscript{\#}   & \textbf{0.917} & \textbf{21.67} & \textbf{0.711} & \textbf{0.878} & \textbf{24.82} & \textbf{0.556}  & 5.803          & 0.816          & 20.6           & \underline{0.329}          & 0.741          \\
WeatherGFM\textsuperscript{\textdagger} & 0.822     & 18.43      & 0.465             & 0.836         & 22.26 & 0.396              & 4.979 & \underline{0.828}          & \underline{21.86}       & 0.325        & 0.777          \\
SynWeatherDiff\textsuperscript{\textdagger}    & 0.868      & 19.79    & \underline{0.690}      & \underline{0.864}  & 23.65  & \underline{0.508}     & \textbf{4.456} & \textbf{0.837} & \textbf{22.33} & \textbf{0.254} & \textbf{0.795} \\

\bottomrule
\end{tabular*}
\caption{Quantitative results on standard weather synthesis tasks. \textsuperscript{\#} specialized model. \textsuperscript{\textdagger}: general model trained with all six weather variable synthesis. Best results are \textbf{bold}, second-best are \underline{underlined}.}
\label{tab:result_standard_tasks}
\end{table*}

\subsection{Network Architecture}
Fig.~\ref{fig:3} outlines the architecture of our general weather variable synthesis model, SynWeatherDiff.
Firstly, it uses a general autoencoder that encodes all weather variables into a shared latent space. And then, a diffusion transformer-based denoising network is trained to perform conditional generation in the latent space, guided by both the encoded satellite inputs and a task-specific text prompt.

\subsubsection{General AutoEncoder. }
Unlike natural images, meteorological images often contain redundancy~\cite{luo2015removing}. It is rare to observe precipitation or typhoons across areas spanning hundreds or thousands of kilometers at the same time. Therefore,  even after careful filtering, meteorological images still include background areas that offer limited meteorological value. 
In addition, some weather variables also exhibit physical similarity and thus have the potential to complement each other.
The above characteristics motivate us to compress various weather variables into a shared latent space. 
Following~\cite{rombach2022high}, we train a general autoencoder using a combination of pixel-wise reconstruction loss, KL divergence loss, and adversarial loss. Specifically, the encoder encodes each weather observation variable $Y_{r,b} \in \mathbb{R}^{1 \times H \times W}$ into latent representation $z_{r, b} \in \mathbb{R}^{C_z \times H_z \times W_z}$ and the decoder reconstructs it as $\hat{Y_{r,b}}$.
 
\subsubsection{Text-Guided Diffusion Transformer. }
During this stage, the diffusion transformer learns to recover clean latent representations of different weather variables from their noisy versions, under the guidance of satellite inputs and task-specific prompts. For each region-variable pair $(r, b)$, we define a prompt $P_{r, b}$ following the format: ``Synthesize the $b$ variable over the $r$ region using corresponding satellite imagery.'' This prompt is embedded using a pretrained CLIP text encoder~\cite{radford2021learning}, with only the final transformer block fine-tuned to adapt to the weather synthesis domain. Similarly, the satellite observation input $X_r$ is encoded by a ViT-based encoder that is jointly trained with the diffusion transformer to extract features. In contrast to SD3~\cite{esser2024scaling}, SynWeatherDiff adopts an early fusion strategy: the noisy latent $z_{r,b}$ is first concatenated with satellite encoder features and then patchified. These patches are further concatenated with the embedded prompt tokens and passed through the self-attention layers of the diffusion transformer to perform conditional denoising.
The training objective in this stage is to predict noise loss as follows:
\begin{equation} 
\mathcal{L} = \mathbb{E}_{z_{r, b}, \epsilon, t} \left[ \left\| \epsilon_\theta(z^t_{r, b}, t, X_{r, b}, P_{r, b}) - \epsilon \right\|_2^2 \right], 
\end{equation} 
\begin{equation} 
z^t_{r, b} = \sqrt{\bar{\alpha}_t} \cdot z_{r, b} + \sqrt{1 - \bar{\alpha}_t} \cdot \epsilon, 
\end{equation} 
where \(\boldsymbol{\epsilon}_\theta(\cdot)\) is the noise predicted by the DiT. $\epsilon \sim \mathcal{N}(0, I)$, and \(\bar{\alpha}_t\) is the cumulative noise schedule product.

\section{Experiments}
\subsection{Evaluation Protocol}
We evaluate performance using three metric categories:

\textbf{Event detection accuracy}: We first use the Critical Success Index (CSI), a common metric in weather variable synthesis~\cite{schaefer1990critical, gong2024cascast}. It is defined as:
\begin{equation}
\text{CSI} = \frac{\text{TP}}{\text{TP} + \text{FP} + \text{FN}},
\end{equation}
where TP, FP, and FN denote the number of true positives, false positives, and false negatives, respectively. Because different variables have different value ranges, we apply task-specific thresholds when computing CSI. For CR synthesis, thresholds of {25, 35, 40} are used, while for precipitation synthesis the thresholds are {2, 5, 15}. For visible light and MWBT, we focus more on perceptual quality. Therefore, we adopt coarse thresholds of 50 and 300, respectively.

\textbf{Regression quality}: In addition to thresholds CSI, we report the Root Mean Squared Error (RMSE) to measure pixel-wise differences between predictions and ground truth. 

\textbf{Perceptual similarity}: Finally, to assess perceptual similarity, we employ the Structural Similarity Index (SSIM), Peak Signal-to-Noise Ratio (PSNR), and Learned Perceptual Image Patch Similarity (LPIPS).

\subsection{Benchmark Setup}
To ensure a comprehensive and fair comparison, we benchmark against three categories of models: (i) general models: WeatherGFM~\cite{zhao2024weathergfm} that focus on multi-variable synthesis but lack regional generalization; (ii) task-specific models: SRViT~\cite{stock2024srvit} for vertically integrated liquid (VIL) and Deep-STEP~\cite{gorooh2022deep}, TomoPE~\cite{cannon2024deep} for precipitation; (iii) popular deep learning models: UNet and ViT adapted for satellite-to-variable mapping. All models are trained using a unified protocol on SynWeather.

\subsection{Implementation Details}
We train SynWeatherDiff for 600K steps with a batch size of 16 on 4×80GB NVIDIA A100 GPUs. The model is optimized using AdamW with a cosine learning rate scheduler, decaying the learning rate from 5e-4 to 1e-5.

\begin{figure*}[ht]
    \centering
    \includegraphics[width=0.80\linewidth]{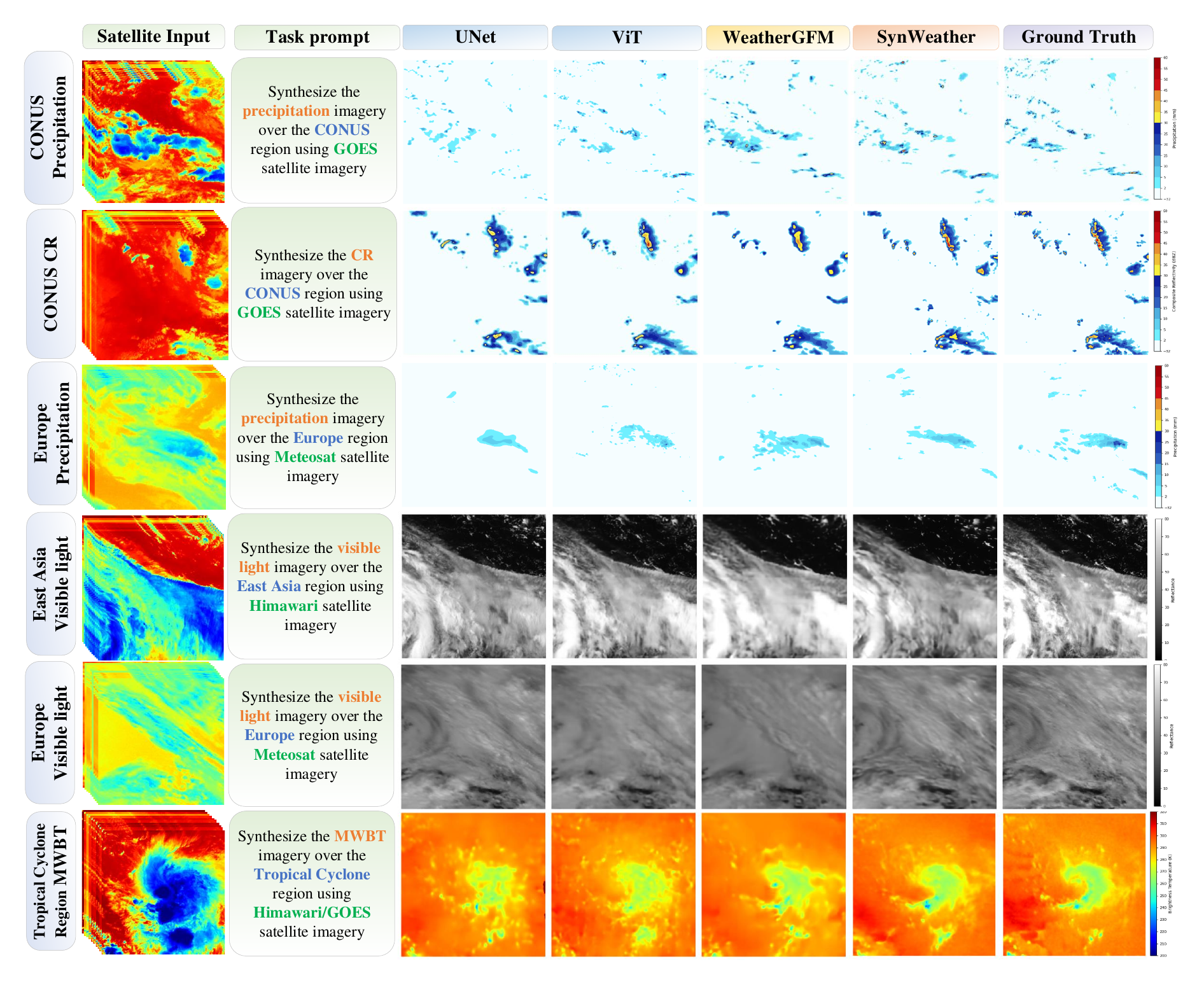}
    \caption{Visual results of the weather synthesis standard tasks by our SynWeatherDiff and other models.}
    \label{fig:6}
\end{figure*}

\begin{table*}[h]
\small
\centering
\begin{tabular}{ccccccccccccc}
\toprule
\multirow{2}{*}{Sampling Setting} & \multicolumn{3}{c}{CONUS Precipitation} & \multicolumn{3}{c}{CONUS CR} & \multicolumn{3}{c}{Europe Visible Light} & \multicolumn{3}{c}{MWBT} \\
\cmidrule(lr){2-4} \cmidrule(lr){5-7} \cmidrule(lr){8-10} \cmidrule(lr){11-13}
& RMSE & CSI/2 & CSI/15 & RMSE & CSI/25 & CSI/35 & SSIM & PSNR & CSI/70 & SSIM & LPIPS & PSNR \\
\midrule Uniform ($\frac{1}{6}$)      & 0.976 & \underline{0.312} & 0.113 & \underline{2.82} & \underline{0.382} & \underline{0.158} & 0.864 & 23.65 & 0.172 & 0.837 & 0.254 & 22.33 \\
CONUS CR ($\frac{1}{2}$) & \textbf{0.961} & \textbf{0.320} & \textbf{0.137} & \textbf{2.69} & \textbf{0.403} & \textbf{0.187} & 0.857 & 23.33 & 0.168 & \textbf{0.843} & \textbf{0.252} & \underline{22.50} \\
CONUS Prec($\frac{1}{2}$)  & \underline{0.974} & 0.292 & 0.087 & 2.88 & 0.343 & 0.110 & 0.855 & 23.34 & 0.166 & 0.841 & 0.258 & 22.15 \\
Europe Vis ($\frac{1}{2}$)  & 1.014 & 0.298 & \underline{0.115} & 2.77 & 0.377 & 0.145 & \underline{0.877} &\textbf{24.16} & \textbf{0.185} & \underline{0.842} & \underline{0.254} & \textbf{22.51} \\
MWBT ($\frac{1}{2}$)   & 1.001 & 0.295 & 0.112 & 2.77 & 0.374 & 0.141 & \textbf{0.879} & \underline{24.13} & \underline{0.184} & \underline{0.842} & \underline{0.254} & 22.38 \\
\bottomrule
\end{tabular}
\caption{Ablation study on task sampling ratios. In each setting, the primary task is assigned a sampling ratio of 0.5, while the remaining five tasks are set to 0.1. We denote each experiment by the name of the primary task along with its sampling ratio. Best results are \textbf{bold}, second-best
are \underline{underlined}. }
\label{tab:result_ablations}
\end{table*}


\subsection{Experiment Results}

\subsubsection{SynWeatherDiff achieves strong universal capabilities. } 
As shown in Table~\ref{tab:result_standard_tasks} and Fig~\ref{fig:6}, SynWeatherDiff, guided by task-specific text prompts,  successfully synthesizes multiple weather variables across different regions using heterogeneous satellite sources. 
It outperforms the existing generalist model, WeatherGFM, across most tasks. 
Furthermore, SynWeatherDiff is capable of distinguishing and generating different variables from the same satellite input, such as synthesizing both CR and precipitation over CONUS, visible light and precipitation over Europe. This highlights our SynWeatherDiff‘s strong universality and flexibility, attributed to the task-specific text prompts.

\subsubsection{SynWeatherDiff outperforms specialized models across most synthesis tasks.} 
As shown in Table~\ref{tab:result_standard_tasks}, SynWeatherDiff achieves state-of-the-art CSI performance with comparable RMSE performance for precipitation synthesis, which is more challenging and critical in meteorological applications. In the visible light synthesis task, UNet achieves better results than SynWeatherDiff as it operates directly in pixel space. Visible light data contains abundant high-frequency details, which are difficult to reconstruct through the autoencoder used by SynWeatherDiff. Nevertheless, with improvements to the autoencoder, SynWeatherDiff holds promising potential to surpass specialized models across all tasks.



\subsubsection{SynWeatherDiff demonstrates a strong ability to generate results with both fine-grained and high-value details.}
In Fig~\ref{fig:6}, SynWeatherDiff clearly outperforms UNet, ViT and WeatherGFM in generating fine-grained details. 
The latter three models produce fewer individual cells, which can be attributed to two factors: (1) weak signals being overlooked; (2) multiple small weather cells tend to merge into larger areas, which smooths out boundaries and removes important local details.
For high-value regions, ViT and WeatherGFM can barely recover some intensity centers in CR but fail to reconstruct the surrounding convective structure. UNet performs worse, often missing the intensity centers entirely in both CR and precipitation tasks. 
In contrast, SynWeatherDiff successfully restores the number and distribution of individual scattered cells. It also accurately captures the location and shape of intensity centers, closely matching the ground truth. These observations are consistent with the quantitative results in Table~\ref{tab:result_ood_tasks}. SynWeatherDiff achieves higher CSI scores and demonstrates superior performance in generating fine-grained patterns and high-intensity events.

\subsection{Ablation Studies and Exploration}
\begin{figure}[t]
    \centering
    \includegraphics[width=1\linewidth]{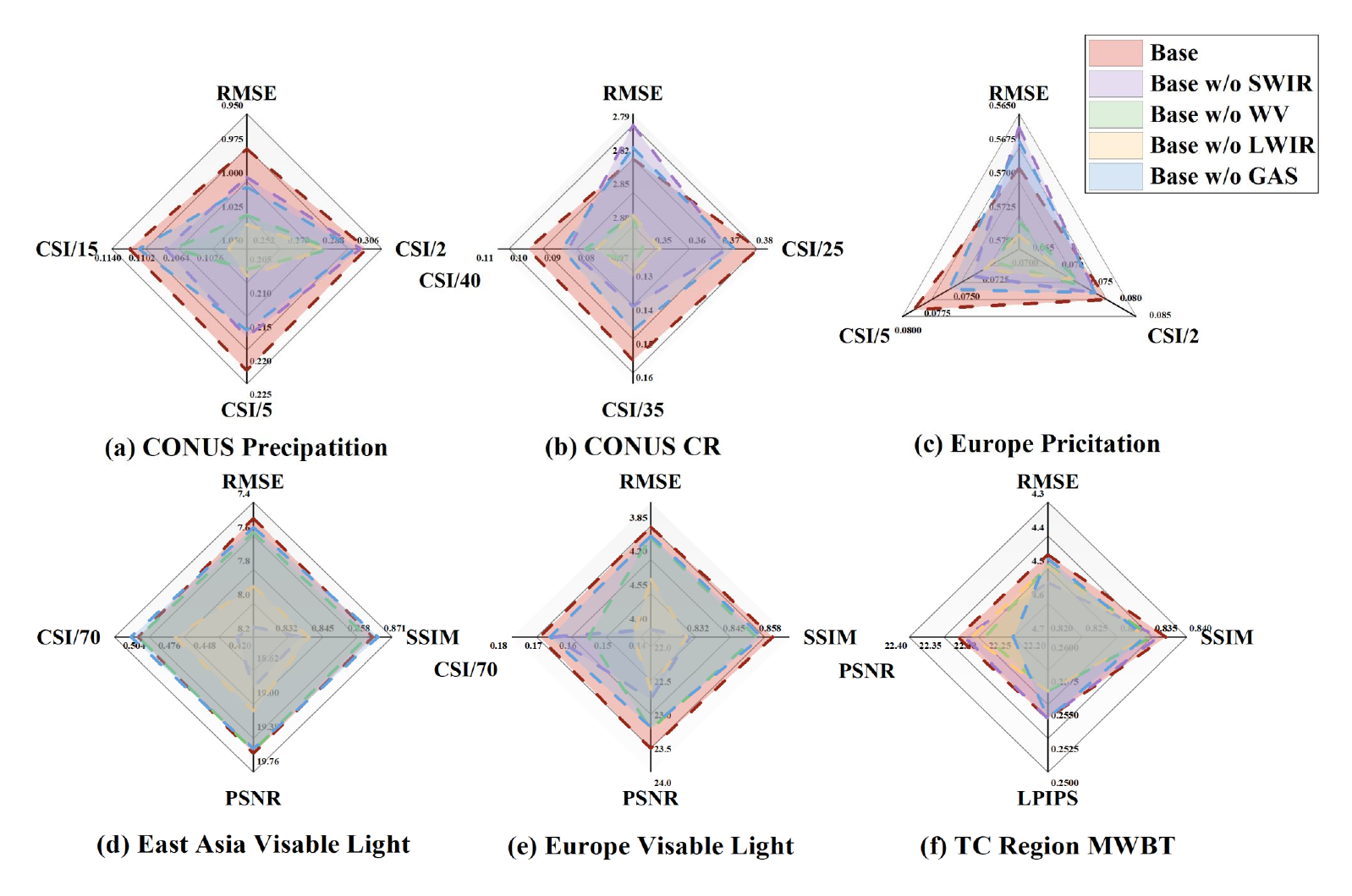}
    \caption{The effect of input channel is analyzed across six tasks, focusing on four groups: (1) SWIR (shortwave infrared), (2) WV (water vapor channels), (3) LWIR (longwave infrared channels), and (4) GAS (gas absorption channels).} 
    \label{fig:5}
\end{figure}

\subsubsection{Exploration of different task sampling ratios.  } 
In this section, CONUS CR, CONUS precipitation, Europe visible light and MWBT synthesis are set as the main task with a 0.5 sampling ratio in turn, with the other five tasks each at 0.1. Table \ref{tab:result_ablations} reveals three findings. (1) CR synthesis helps CONUS precipitation (as the Z-R relationship makes CR a strong prior, and its large sample size and convective pattern capturing ability benefit precipitation modeling ) and MWBT synthesis (since CR from ground-based S-band radars shares a similar physical domain with satellite-based sensors and offers higher spatial resolution ). Thus, a higher CR task sampling ratio than uniform sampling boosts both. (2) Visible light and MWBT tasks benefit each other. When Europe's visible light or tropical cyclone MWBT synthesis is the main task, they surpass uniform sampling. This is because the visible band and MWBT have spatial correspondence in strong convective systems, aiding cross-task knowledge transfer. (3) CR task performance gains mainly come from the increased number of training samples. It does not benefit from interactions with other tasks. Notably, the CR task only improves when CR synthesis is set as the main task. Also, there are conflicting tasks like CR and visible light synthesis. Therefore, choosing a proper task ratio to balance mutually beneficial and conflicting tasks is key for overall performance gains.

\subsubsection{Exploration of different input channels. } 
We choose all 10 infrared channels instead of the usual 3-channel setup, and divided them into four groups based on spectral range and atmospheric sensitivity (SWIR, WV, LWIR, GAS). Ablation experiments involved removing each group to assess its contribution to synthesis performance. Fig. \ref{fig:5} shows that removing any group worsens all synthesis tasks. Detailed analysis indicates that water vapor and longwave infrared channels are vital for precipitation and radar reflectivity synthesis, while shortwave infrared channels are more important for visible light synthesis. For example, in the East Asia visible light task, removing the SWIR channel alone causes a bigger performance drop than removing all LWIR channels. Gas absorption channels have less impact on visible light synthesis. For MWBT synthesis, removing any single group reduces performance, but the overall effect is small.



\begin{table}[ht]
\small
\centering
\begin{tabular}{lccccc}
\toprule
Metric & RMSE & SSIM & PSNR & CSI/2 & CSI/5 \\
\midrule
UNet\textsuperscript{\#}    & 1.383 & 0.671 & \textbf{30.02} & 0.161 & 0.076 \\
ViT\textsuperscript{\#}     & \underline{1.233} & \underline{0.749} & 29.49 & \underline{0.232} & \underline{0.103} \\
SynWeatherDiff\textsuperscript{\textdagger} & \textbf{1.150} & \textbf{0.771} &  \underline{29.80} & \textbf{0.235} & \textbf{0.108} \\
\bottomrule
\end{tabular}
\caption{Quantitative results on OOD weather synthesis tasks. \textsuperscript{\#} specialized model. \textsuperscript{\textdagger}: general model. Best results are \textbf{bold}, and second-best are \underline{underlined}.}
\label{tab:result_ood_tasks}
\end{table}

\subsubsection{Exploration on out-of-distribution tasks. } 
To evaluate the generalization ability of SynWeatherDiff, we conduct OOD experiments on East Asia precipitation synthesis. 
Table~\ref{tab:result_ood_tasks} shows that SynWeatherDiff surpasses the specialized model trained exclusively on CONUS precipitation, demonstrating superior generalization across both region and variable.
This not only highlights the advantages of using text prompts for flexible task control but also shows that different variables can complement each other in a general model, compared to region-specific and single-variable models.
Nevertheless, inference under the current prompt-based framework is also constrained to the specific regions and variables that were individually present in the training set.


\section{Conclusion}
We construct the first multi-region and multi-variable weather observation data synthesis dataset (i.e., SynWeather) and introduce a general diffusion-based model as a baseline (i.e., SynWeatherDiff). By introducing text prompts to guide the synthesis process, SynWeatherDiff flexibly generates diverse weather variables across multiple regions within a unified framework. We conduct a comprehensive analysis of task sampling ratios and the input channel. Through the proposed dataset and baseline model, we aim to facilitate research in weather synthesis and inspire the development of future generalist models for weather-related downstream tasks.

\section{Acknowledgements}
This work is supported by Shanghai Artificial Intelligence Laboratory. This work was done during her internship at Shanghai Artificial Intelligence Laboratory.

\bibliography{aaai2026}

\newpage
\newpage

\section{A. More SynWeather Dataset Details}
This section provides additional details about SynWeather, including raw data acquisition, standardized preprocessing procedures, and the strategy for training, validation, and test splits.
\subsection{A1. Raw Data Acquisition and Preprocessing. }
For the satellite inputs, we collect infrared brightness temperature observations from three major geostationary satellite series. The specific sources are as follows:
\begin{itemize}
    \item \textbf{GOES-16/17/18:} ABI L1b RadC products are obtained from NOAA S3\footnote{\url{https://noaa-goes16.s3.amazonaws.com/index.html#}}%
    \footnote{\url{https://noaa-goes17.s3.amazonaws.com/index.html#}}%
    \footnote{\url{https://noaa-goes18.s3.amazonaws.com/index.html#}}.
    
    \item \textbf{Himawari-8/9:} AHI L1b FLDK data are also provided by NOAA\footnote{\url{https://noaa-himawari8.s3.amazonaws.com/index.html#}}.
    
    \item \textbf{Meteosat-11:} SEVIRI Level 1.5 products are downloaded from the EUMETSAT Data Store\footnote{\url{https://data.eumetsat.int}}.
\end{itemize}

All raw satellite observation files are processed using the SatPy library~\cite{hoese2019satpy}.
We first define the target regions and select satellite data based on spatial coverage to ensure consistency between geostationary inputs and the corresponding meteorological targets.

To unify spatial representation, we use SatPy’s \texttt{resample} function to project all satellite observations onto a uniform grid based on the equidistant cylindrical projection (WGS84 ellipsoid).
This resampling standardizes spatial layout and facilitates alignment across multiple data sources.
Although the native resolutions vary (2 km for GOES and Himawari, and 3 km for Meteosat), observations are resampled to a consistent 4 km resolution.
This choice balances computational efficiency and ensures compatibility across satellite sources.
Table~\ref{tab:region-definition} summarizes the coverage and grid specifications for each region.

To enhance input quality, SynWeather adopts a richer channel configuration than prior datasets. Specifically:

\begin{itemize}
    \item \textbf{GOES-16/17/18 and Himawari-8/9:} We use all infrared channels (C07–C16), which cover shortwave, water vapor, and longwave infrared bands. These channels are essential for fire detection, moisture tracking, and surface temperature retrieval.

    \item \textbf{Meteosat-11:} As it has fewer infrared channels compared to GOES and Himawari, we replicate bands with similar central wavelengths where necessary. This ensures consistent input dimensionality across different satellite sources.
\end{itemize}

\begin{table*}[ht]
\centering
\caption{Spatial coverage and grid configuration of each region. }
\label{tab:region-definition}
\begin{tabular}{ccccc}
\toprule
\textbf{Region} & \textbf{Latitude Range} & \textbf{Longitude Range} & \textbf{Spatial Resolution} & \textbf{Grid Size} \\
\midrule
CONUS     & 25°N–47°N   & 74°W–121°W   & 4 km & (550, 1175)  \\
East Asia & 4°N–53°N    & 73°E–150°E   & 4 km & (1225, 1925) \\
Europe    & 32°N–68°N   & 40°W–58°E    & 4 km & (900, 2450)  \\
\bottomrule
\end{tabular}
\end{table*}

For weather variables, we collect observation data from the following sources:
\begin{itemize}
    \item \textbf{CONUS Composite Reflectivity (CR):} Retrieved from the Dryad repository\footnote{\url{https://datadryad.org/dataset/doi:10.5061/dryad.h9w0vt4nq}}%
    \footnote{\url{https://datadryad.org/dataset/doi:10.5061/dryad.zs7h44jf2}}%
    \footnote{\url{https://datadryad.org/dataset/doi:10.5061/dryad.2jm63xstt}}.

    \item \textbf{CONUS Precipitation:} Retrieved from the Iowa State Mesonet Archive\footnote{\url{https://mtarchive.geol.iastate.edu}}.

    \item \textbf{East Asia and Europe Visible Light:} 
    Himawari-8 AHI L1b FLDK data\footnote{\url{https://noaa-himawari8.s3.amazonaws.com/index.html#AHI-L1b-FLDK/}} for East Asia and Meteosat-11 SEVIRI Level 1.5 products from EUMETSAT\footnote{\url{https://data.eumetsat.int}} for Europe. 
    Since the visible channels differ across satellites, we choose the closest spectral equivalents: Himawari-8 C02 and Meteosat-11 VIS006.

    \item \textbf{East Asia Precipitation:} Obtained from the GPM Mission\footnote{\url{https://gpm.nasa.gov/missions/GPM}}.

    \item \textbf{Europe Precipitation:} Obtained from the EURADCLIM dataset\footnote{\url{https://dataplatform.knmi.nl/dataset/rad-opera-hourly-rainfall-accumulation-euradclim-2-0}}.
\end{itemize}

All target variables are first reprojected into the equidistant cylindrical projection and resampled to a uniform spatial resolution of 4\,km, consistent with the satellite inputs.

As the spatial extent varies across different region-variable pairs as shown in Table~\ref{tab:region-definition}, we crop all variables into fixed-size patches of $256 \times 256$ using a sliding window with a stride of $128$ pixels. To filter out background-dominated patches, we define two thresholds $\gamma_1$ and $\gamma_2$, and retain only those patches that contain connected components formed by pixels exceeding $\gamma_1$ in value, with spatial areas larger than $\gamma_2$. Note that microwave brightness temperature (MWBT) data is organized by tropical cyclone (TC) events and thus is not subject to the above cropping and filtering process. The specific threshold settings are summarized in Table~\ref{tab:patch-thresholds}.
\begin{table}[t]
\centering
\caption{Thresholds used for patch selection. $\gamma_1$ is the intensity threshold, and $\gamma_2$ is the minimum area of connected components.}
\label{tab:patch-thresholds}
\begin{tabular}{ccc}
\toprule
\textbf{Variable} & $\boldsymbol{\gamma_1}$ & $\boldsymbol{\gamma_2}$ \\
\midrule
CR & 8  & 600  \\
Precipitation               & 5  & 500  \\
Visible Light               & 30 & 1200 \\
\bottomrule
\end{tabular}
\end{table}

\subsection{A2. SynWeather Dataset Split. }
For efficient training, we use filtered image patches as input instead of full-region images. Since the ultimate goal is to generate meteorological variables at the regional scale, evaluation during inference is still performed on the complete image. 
Specifically, each regional image is divided into $256 \times 256$ patches, which are individually processed by the model. 
The outputs are then reassembled to reconstruct the full image, upon which evaluation metrics are computed.

Considering the practical value of accurately synthesizing heavy rainfall and reflectivity, we use three fixed time points per day from July in the final year of the dataset as our test set for precipitation, CR, and visible light tasks. The remaining filtered patches from July are used as the validation set. The number of samples for each split is summarized in Table~\ref{tab:sample-split}.

\begin{table}[t]
\centering
\caption{Sample numbers for each variable and region. Visible light has significantly more training samples due to its wider spatial coverage and higher information density. Precipitation data is inherently sparse, and strict filtering further reduces its sample count. MWBT data has the fewest samples, primarily due to the limited temporal coverage of polar-orbiting satellites and the rarity of extreme typhoon events. For the test set, except for MWBT data, patches are extracted from full-region maps at selected timestamps without any filtering, resulting in consistent sample counts across variables within the same region. }
\begin{tabular}{ccccc}
\toprule
\multirow{2}{*}{\textbf{Variable}} & \multirow{2}{*}{\textbf{Region}} & \multicolumn{3}{c}{\textbf{Sample number}} \\
\cmidrule(lr){3-5}
                  &                 & \textbf{Train} & \textbf{Valid} & \textbf{Test} \\
\midrule
\multirow{2}{*}{Visible light} 
                  & Europe          & 346k         & 11k          & 14k        \\
                  & East Asia       & 478k        & 9k           & 14k        \\
\midrule
CR                & CONUS           & 13k          & 7k          & 4k         \\
\midrule
\multirow{2}{*}{Precipitation} 
                  & Europe          & 9k          & 768            & 14k         \\
                  & CONUS           & 14k          & 707            & 4k         \\
\midrule
MWBT         & TC Region       & 7k           & 1k           & 976           \\
\bottomrule
\end{tabular}
\label{tab:sample-split}
\end{table}

\section{B. Baseline Models for Comparison}
We benchmark SynWeatherDiff against a diverse set of representative baselines covering both generalist and specialized approaches.
For general methods, we include WeatherGFM~\cite{zhao2024weathergfm}, a large-scale weather foundation model trained across a wide range of weather understanding tasks using visual in-context prompts.
For specialized methods, we evaluate SRViT~\cite{stock2024srvit}, a transformer-based architecture optimized with weighted loss to better capture rare but high-intensity radar reflectivity values in CR synthesis;
Deep-STEP~\cite{gorooh2022deep}, which estimates precipitation at 4 km resolution by fusing microwave and infrared brightness temperatures; and
TomoPE~\cite{cannon2024deep}, which utilizes the EfficientNetV2~\cite{tan2021efficientnetv2} architecture to predict precipitation rates from combined geostationary and low-Earth-orbit satellite observations.
In addition to these domain-specific models, we also adapt popular deep learning architectures, (i.e., ViT~\cite{dosovitskiy2020image} and UNet~\cite{ronneberger2015u}) for satellite-to-weather variable mapping. These serve as strong baselines, all trained under a consistent experimental protocol.

\begin{table*}[ht]
\centering
\caption{Hyperparameters of SynWeatherDiff.}
\label{tab:hyperparams}
\begin{tabular}{cllc}
\toprule
\textbf{Module} & \textbf{Hyperparameter} & \textbf{Meaning} & \textbf{Value} \\
\midrule
\multirow{8}{*}{Autoencoder}
  & Input size & Size of input image & 256 \\
  & Input channels & Number of input channels & 1 \\
  & Output channels & Number of output channels & 1 \\
  & Downsampling ratio & Spatial downsampling ratio & 8$\times$ \\
  & Layers per block & Number of layers per block & 2 \\
  & Latent channels & Channels in latent representation & 4 \\
  & Group norm groups & Number of groups for GroupNorm & 32 \\
\midrule
\multirow{10}{*}{DiT}
  & Input size & Spatial size of latent input & 32 \\
  & Input channels & Number of input feature channels & 14 \\
  & MLP ratio & Expansion ratio in MLP & 4.0 \\
  & Output channels & Output latent channels & 4 \\
  & Attention heads & Number of attention heads & 16 \\
  & $p_l$ & Patch size in latent space & 2 \\
  & Latent depth & Transformer blocks in latent stage & 12 \\
  & Encoder depth & Transformer blocks in encoder & 12 \\
  & Hidden size & Transformer hidden size & 512 \\
  & Encoder hidden size & Hidden size in encoder & 256 \\
\midrule
\multirow{6}{*}{Diffusion}
  & Timesteps & Number of training steps & 1000 \\
  & Beta schedule & Variance schedule type & Linear \\
  & Beta start & Starting beta value & 0.0001 \\
  & Beta end & Ending beta value & 0.02 \\
  & Sampling method & Sampling strategy used during inference & DDIM \\
  & Sampling steps & Number of sampling steps & 20 \\
\bottomrule
\end{tabular}
\end{table*}

\section{C. Implementation Details}
\subsection{C1. Implementation details and hyperparameters of SynWeatherDiff}
The hyperparameters for SynWeatherDiff is shown in Table~\ref{tab:hyperparams}.
We train SynWeatherDiff for 600K steps with a batch size of 16 on 4$\times$80GB NVIDIA A100 GPUs. The model is optimized using AdamW with a cosine learning rate scheduler, decaying the learning rate from $5\times10^{-4}$ to $1\times10^{-5}$. A warm-up phase is applied to stabilize early training. The optimizer uses $\beta_1 = 0.9$ and $\beta_2 = 0.95$ to better adapt to the dynamics of diffusion-based training.

\subsection{C2. WeatherGFM Hyperparameters}
We adopt the WeatherGFM implementation provided by~\cite{zhao2024weathergfm}. The hyperparameters listed below are consistently used across all WeatherGFM experiments.

\begin{table}[ht]
\centering
\scriptsize
\caption{Hyperparameters of WeatherGFM.}
\label{tab:weathergfm-hyperparams}
\begin{tabular}{lll}
\toprule
\textbf{Hyperparameter} & \textbf{Description} & \textbf{Base} \\
\midrule
$p$                     & Patch size                                & 16   \\
Encoder dimension       & Dimensionality of encoder embeddings       & 768  \\
Decoder dimension       & Dimensionality of decoder embeddings       & 512  \\
Encoder depth           & Number of transformer blocks in encoder    & 12   \\
Decoder depth           & Number of transformer blocks in decoder    & 8    \\
Encoder heads           & Number of self-attention heads in encoder  & 12   \\
Decoder heads           & Number of self-attention heads in decoder  & 16   \\
MLP ratio               & Expansion ratio in MLP layers              & 4    \\
Masked ratio            & Proportion of target tokens being masked   & 75\% \\
\bottomrule
\end{tabular}
\end{table}

\section{D. Experiments on Few-Shot tasks}

\begin{figure}[!h]
    \centering
    \includegraphics[width=0.9\linewidth]{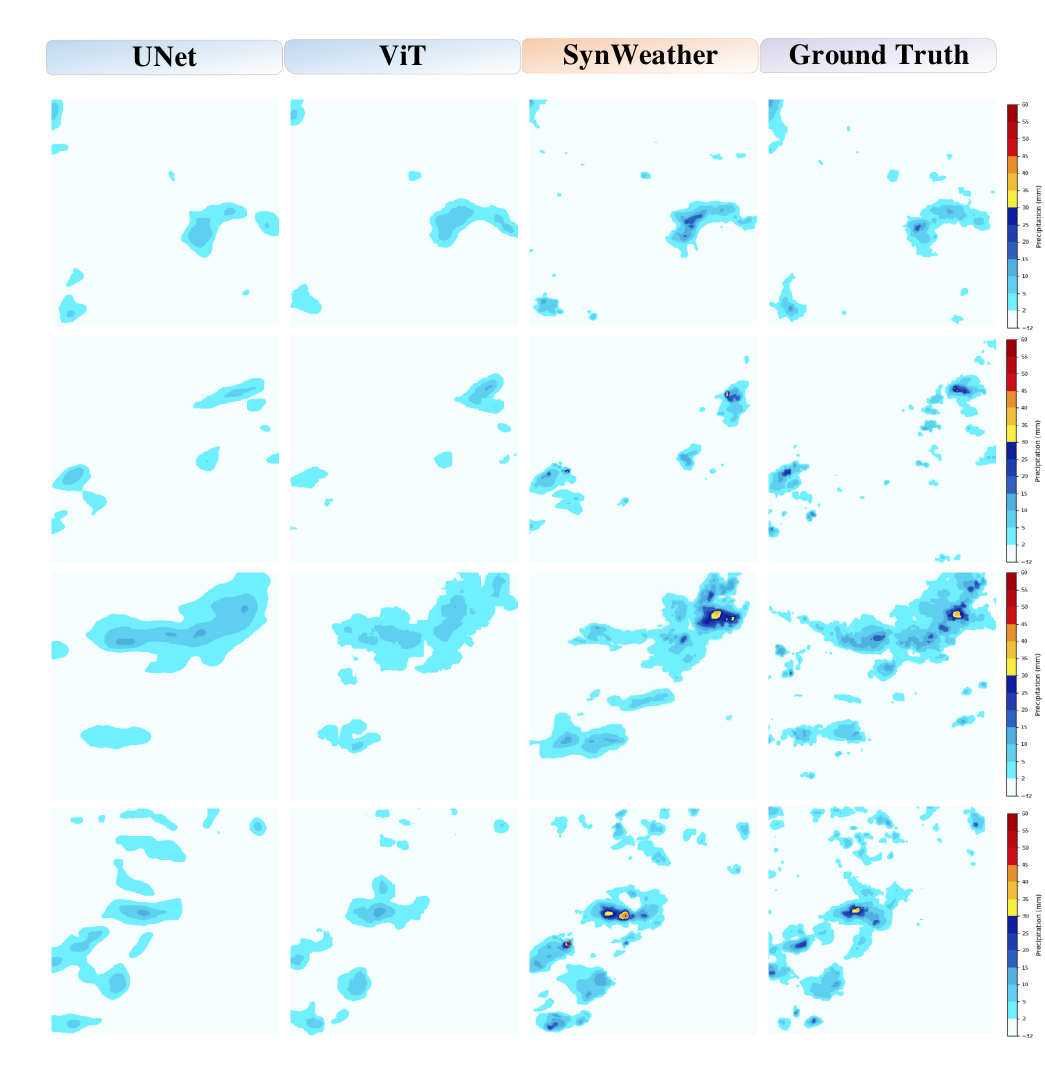}
    \caption{Visual results of the weather synthesis on Few-Shot tasks.}
    \label{fig:ood}
\end{figure}

To evaluate the cross-region adaptability of weather synthesis models, we perform few-shot experiments on the East Asia precipitation synthesis task. 
Specifically, we fine-tune the general model SynWeatherDiff and the specialized models ViT and UNet, all of which are pre-trained on CONUS precipitation data. 
Each model is trained for 1000 steps (approximately 2.7 epochs) using the AdamW optimizer with a cosine learning rate scheduler, decaying the learning rate from $5\times10^{-4}$ to $1\times10^{-5}$. 
As shown in Table~\ref{result_few_shot_tasks}, SynWeatherDiff achieves the best performance in SSIM and PSNR, indicating strong capabilities in structural reconstruction and pixel-level fidelity. Although its CSI/2 and CSI/5 scores at the POOL1 level are slightly lower than those of UNet and ViT, this is primarily due to the ground truth being interpolated from 10km GPM data to 4km resolution. 
The smoothing effect which is introduced during interpolation weakens the evaluation of fine-grained details generated by the model. 
However, as the pooling scale increases from POOL1 to POOL8, SynWeatherDiff shows a significant rise in CSI scores, with a much steeper growth trend than other models, ultimately achieving the highest CSI/5 and CSI/15 at POOL8. 
This trend demonstrates that SynWeatherDiff effectively preserves high-resolution details learned from 4km satellite inputs, and its structural advantages become more apparent at coarser evaluation levels. 
These results further confirm that SynWeatherDiff excels not only in capturing large-scale spatial distributions but also in reconstructing localized extreme precipitation events, consistent with the visual results in Fig.~\ref{fig:ood}.

\begin{table*}[ht]
\centering
\scriptsize
\caption{Quantitative results on few shot tasks. \textsuperscript{\#} specialized model. \textsuperscript{\textdagger}: general model trained with all six weather variable synthesis. Best results are \textbf{bold}, second-best are \underline{underlined}.}
\begin{tabular}{lcccccccccccc}
\toprule
\multirow{2}{*}{Metric} & \multirow{2}{*}{SSIM} & \multirow{2}{*}{PSNR} 
& \multicolumn{3}{c}{CSI/2} & \multicolumn{3}{c}{CSI/5} & \multicolumn{3}{c}{CSI/15} \\
\cmidrule(lr){4-6} \cmidrule(lr){7-9} \cmidrule(lr){10-12}
& & & POOL1 & POOL4 & POOL8 & POOL1 & POOL4 & POOL8 & POOL1 & POOL4 & POOL8 \\
\midrule
ViT\textsuperscript{\#}            & \underline{0.791} & \underline{29.86} & \underline{0.411} & \underline{0.418} & 0.425 & \underline{0.278} & 0.279 & 0.281 & \textbf{0.061} & \underline{0.060} & \underline{0.056} \\
UNet\textsuperscript{\#}          & 0.431 & 27.94 & \textbf{0.430} & \textbf{0.441} & \textbf{0.454} & \textbf{0.290} & \textbf{0.296} & \underline{0.305} & 0.042 & 0.040 & 0.041 \\
SynWeatherDiff\textsuperscript{\textdagger} & \textbf{0.828} & \textbf{29.97} & 0.373 & 0.402 & \underline{0.436} & 0.252 & \underline{0.281} & \textbf{0.314} & \underline{0.048} & \textbf{0.064} & \textbf{0.081} \\
\bottomrule
\end{tabular}
\label{result_few_shot_tasks}
\end{table*}

\begin{table*}[ht]
\centering
\caption{POOL-X CSI scores on Precipitation synthesis. \textsuperscript{\#} specialized model. \textsuperscript{\textdagger}: general model trained with all six weather variable synthesis. Best results are \textbf{bold}, second-best are \underline{underlined}.}
\begin{tabular}{lcccccccccc}
\toprule
\textbf{Region} & \multicolumn{6}{c}{\textbf{CONUS}} & \multicolumn{4}{c}{\textbf{Europe}} \\
\cmidrule(lr){2-7} \cmidrule(lr){8-11}
\multirow{2}{*}{\textbf{Metric}}  & \multicolumn{3}{c}{POOL4}  & \multicolumn{3}{c}{POOL8} & \multicolumn{2}{c}{POOL4} & \multicolumn{2}{c}{POOL8} \\
\cmidrule(lr){2-4} \cmidrule(lr){5-7} \cmidrule(lr){8-9} \cmidrule(lr){10-11}
 & CSI/2 & CSI/5 & CSI/15 & CSI/2 & CSI/5 & CSI/15 & CSI/2 & CSI/5 & CSI/2 & CSI/5 \\
\midrule
Deep-STEP\textsuperscript{\#}        & 0.298 & 0.110 & 0.009  & 0.326 & 0.120 & 0.011  & \underline{0.092} & 0.018 & 0.100 & 0.020 \\
TomoPE\textsuperscript{\#}           & 0.307 & 0.199 & 0.067  & 0.357 & 0.251 & 0.108  & 0.069 & 0.014 & \underline{0.080} & 0.020 \\
UNet\textsuperscript{\#}             & 0.299 & 0.237 & 0.116  & 0.363 & 0.300 & 0.171  & 0.082 & 0.042 & 0.130 & \underline{0.072} \\
ViT\textsuperscript{\#}              & 0.319 & 0.229 & 0.126  & 0.376 & 0.281 & 0.164  & 0.110 & \underline{0.083} & 0.153 & 0.114 \\
WeatherGFM\textsuperscript{\textdagger}      & \underline{0.374} & \underline{0.290} & \underline{0.170}  & \underline{0.441} & \underline{0.363} & \underline{0.233}  & 0.036 & 0.033 & 0.058 & 0.053 \\
SynWeatherDiff\textsuperscript{\textdagger}  & \textbf{0.408} & \textbf{0.321} & \textbf{0.192}  & \textbf{0.480} & \textbf{0.394} & \textbf{0.259}  & \textbf{0.119} & \textbf{0.112} & \textbf{0.144} & \textbf{0.139} \\
\bottomrule
\end{tabular}
\label{pool_Prec_synthesis}
\end{table*}

\begin{table*}[ht]
\centering
\caption{POOL-X CSI scores on CR synthesis. \textsuperscript{\#} specialized model. \textsuperscript{\textdagger}: general model trained with all six weather variable synthesis. Best results are \textbf{bold}, second-best are \underline{underlined}.}
\begin{tabular}{lcccccc}
\toprule
\multirow{2}{*}{\textbf{Metric}} & \multicolumn{3}{c}{\textbf{POOL4}} & \multicolumn{3}{c}{\textbf{POOL8}} \\
\cmidrule(lr){2-4} \cmidrule(lr){5-7}
 & CSI/25 & CSI/35 & CSI/40 & CSI/25 & CSI/35 & CSI/40 \\
\midrule
SRViT\textsuperscript{\#}           & 0.329  & 0.194  & \underline{0.134}  & 0.376 & 0.228  & \underline{0.153}  \\
UNet\textsuperscript{\#}            & 0.360  & 0.135  & 0.071  & 0.416  & 0.193  & 0.123  \\
ViT\textsuperscript{\#}             & 0.383  & \underline{0.209}  & 0.120  & \underline{0.459}  & \underline{0.234}  & 0.146  \\
WeatherGFM\textsuperscript{\textdagger}     & \underline{0.395}  & 0.193  & 0.099 & 0.420  & 0.215  & 0.112 \\
SynWeatherDiff\textsuperscript{\textdagger} & \textbf{0.436}  & \textbf{0.221}  & \textbf{0.144}  & \textbf{0.477}  & \textbf{0.270}  & \textbf{0.173}  \\
\bottomrule
\end{tabular}
\label{pool_CR_synthesis}
\end{table*}

\section{E. Additional Quantitative Results}
In this section, we present additional quantitative results for weather variable synthesis tasks.
Table~\ref{pool_Prec_synthesis} and Table~\ref{pool_CR_synthesis} report the POOL-X CSI scores for precipitation and CR synthesis, which are widely adopted in the meteorological community~\cite{he2025diffsr, gong2024postcast}.
The POOL-X operation refers to applying a max pooling with kernel size and stride of X to both the prediction and the ground truth before computing the CSI.
This technique highlights coarse-grained spatial structures and mitigates the penalization caused by small-scale spatial misalignments, making it particularly suitable for evaluating meteorological predictions under spatial uncertainty.
As shown in the results, SynWeatherDiff consistently achieves state-of-the-art performance across both CR and precipitation synthesis tasks.

\section{F. More Visual Results}
\begin{figure*}[ht]
    \centering
    \includegraphics[width=1\linewidth]{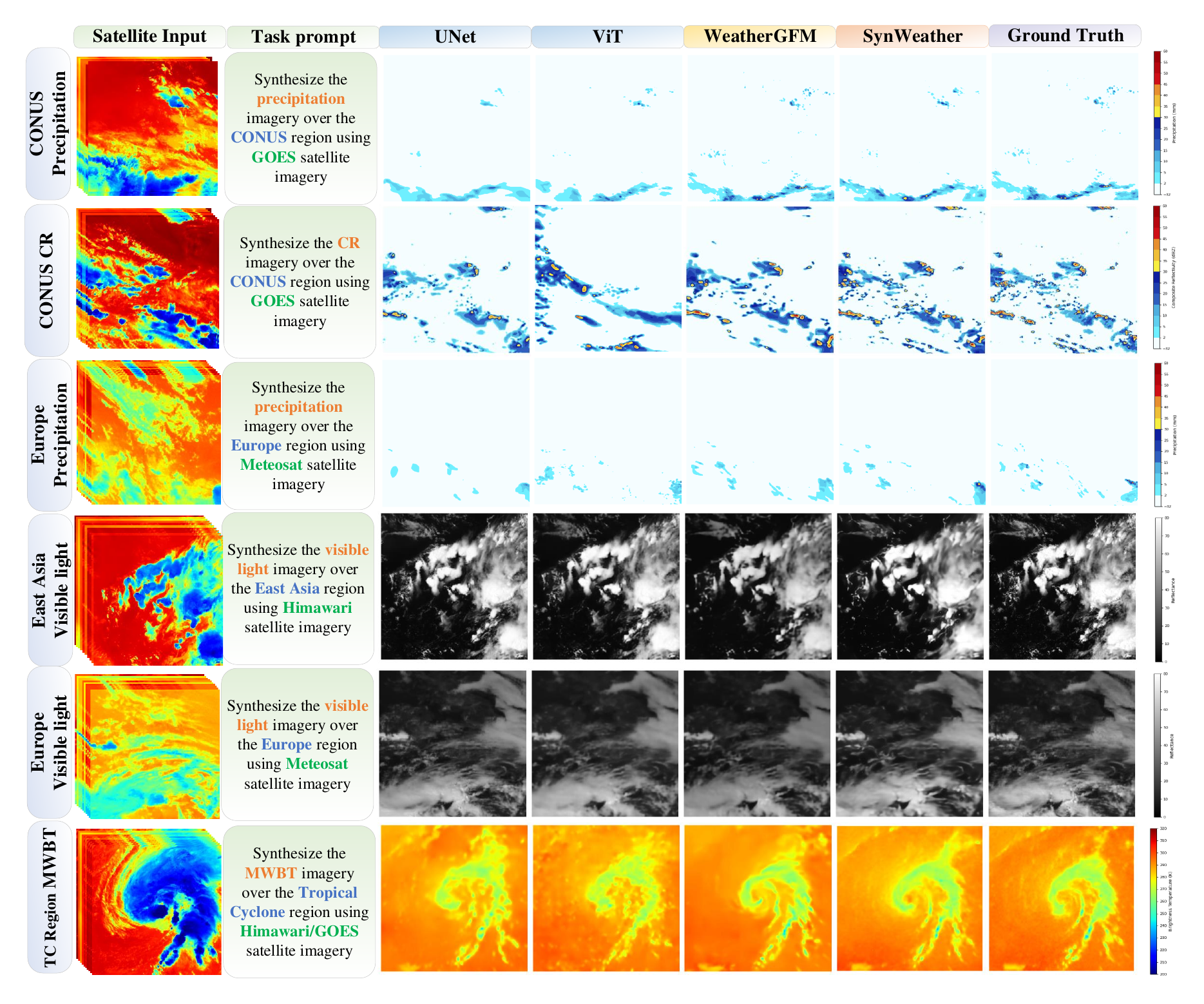}
    \caption{Visual results of the weather synthesis standard tasks by our SynWeatherDiff and other models.}
    \label{fig:r1}
\end{figure*}

\begin{figure*}[ht]
    \centering
    \includegraphics[width=1\linewidth]{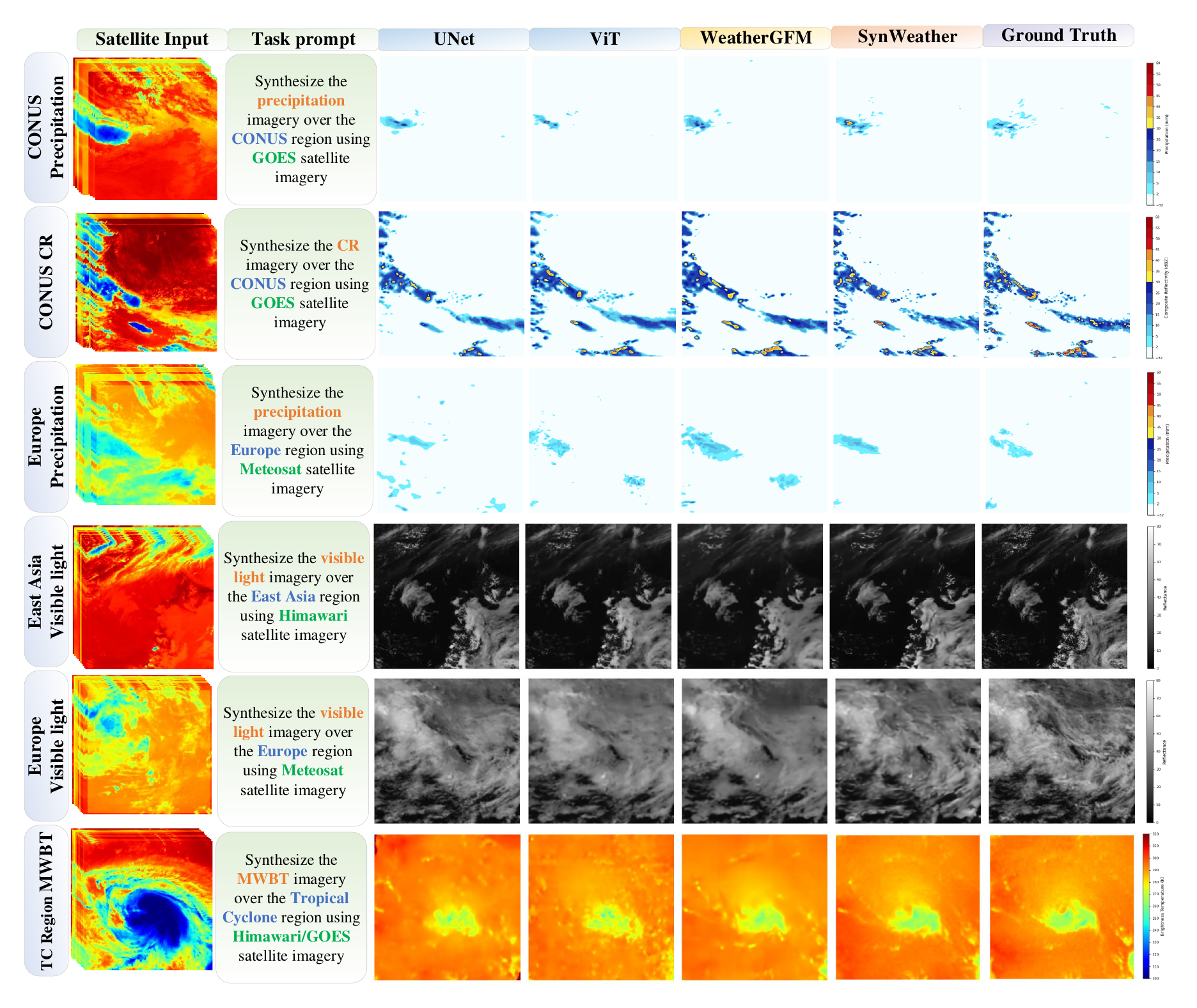}
    \caption{Visual results of the weather synthesis standard tasks by our SynWeatherDiff and other models.}
    \label{fig:r2}
\end{figure*}

\begin{figure*}[ht]
    \centering
    \includegraphics[width=1\linewidth]{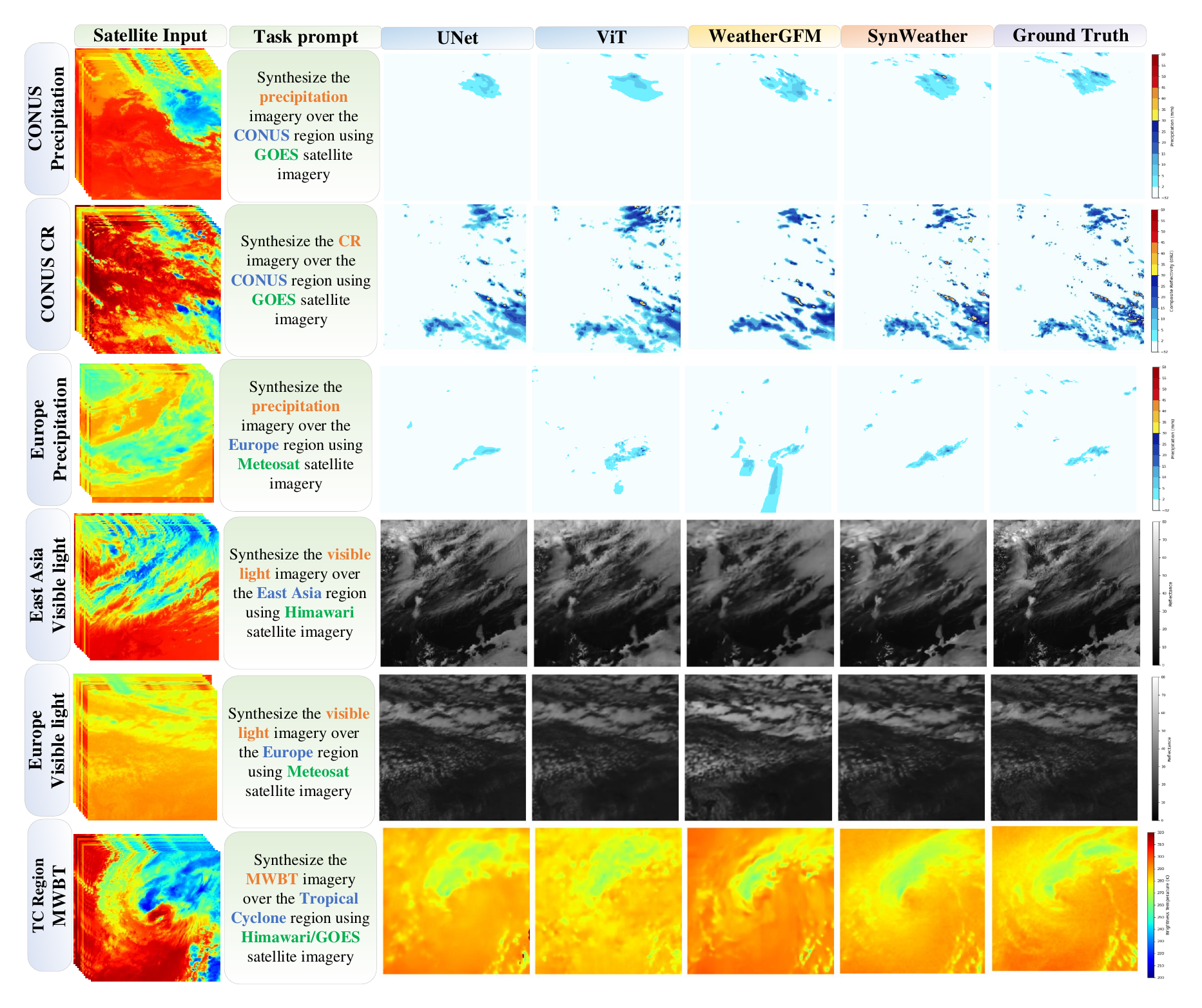}
    \caption{Visual results of the weather synthesis standard tasks by our SynWeatherDiff and other models.}
    \label{fig:r3}
\end{figure*}

\begin{figure*}[ht]
    \centering
    \includegraphics[width=1\linewidth]{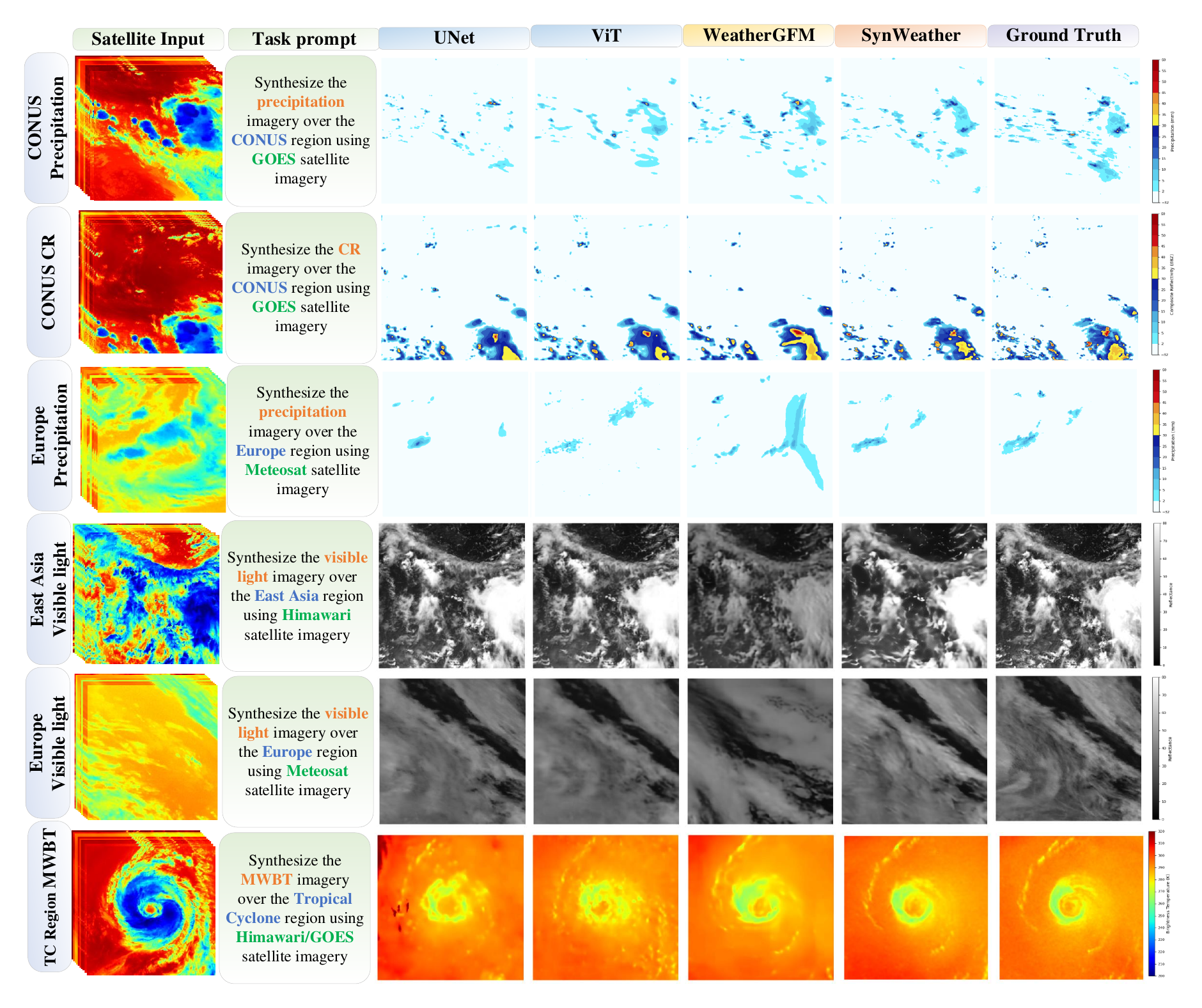}
    \caption{Visual results of the weather synthesis standard tasks by our SynWeatherDiff and other models.}
    \label{fig:r4}
\end{figure*}

To intuitively demonstrate the effectiveness of SynWeatherDiff, we present a series of qualitative visualizations across six synthesis tasks, including CR and precipitation over the CONUS region, visible light and precipitation over Europe, visible light over East Asia, and microwave brightness temperature over tropical cyclone regions. 
We also conduct comparative evaluations against specialized models (i.e., UNet and ViT) as well as the general-purpose model WeatherGFM. The corresponding visual results and comparisons are illustrated in Figure~\ref{fig:r1}, Figure~\ref{fig:r2}, Figure~\ref{fig:r3}, and Figure~\ref{fig:r4}.
The results demonstrate that SynWeatherDiff consistently outperforms all baseline models in both fine-grained structural reconstruction and high-value region synthesis. Notably, SynWeatherDiff leverages both text prompts and satellite observations to guide the generation process. 
This enables flexible task control in various complex scenarios. The dual-input design helps the model go beyond the limitations of traditional models that rely on fixed tasks or regions. As a result, SynWeatherDiff serves as a truly general-purpose weather synthesis model and consistently delivers strong performance across diverse applications.

\FloatBarrier



\end{document}